\documentclass[final]{cvpr}

\usepackage[pagebackref=true,breaklinks=true,colorlinks,bookmarks=false]{hyperref}

\def\cvprPaperID{2782} %
\def\confYear{CVPR 2021}

\usepackage[utf8]{inputenc} %
\usepackage[T1]{fontenc}    %
\usepackage{booktabs}       %
\usepackage{nicefrac}       %
\usepackage{microtype}      %

\usepackage{times}
\usepackage{soul}

\usepackage{epsfig}
\usepackage{amsmath}
\usepackage{amsfonts}
\usepackage{graphics}
\usepackage{graphicx}
\usepackage{color}
\usepackage{bbm}
\usepackage{graphicx}
\usepackage{wrapfig}
\usepackage{tabularx}
\usepackage{caption}
\usepackage[normalem]{ulem}
\usepackage{hyperref}       %
\usepackage{url}            %
\usepackage{xcolor}
\usepackage{bbm}

\usepackage{amsmath,amsfonts,bm}

\def\1{\bm{1}}

\newcommand{\test}{\mathcal{D_{\mathrm{test}}}}

\DeclareMathAlphabet{\mathsfit}{\encodingdefault}{\sfdefault}{m}{sl}
\SetMathAlphabet{\mathsfit}{bold}{\encodingdefault}{\sfdefault}{bx}{n}

\newcommand{\E}{\mathbb{E}}

\newcommand{\R}{\mathbb{R}}

\usepackage{url}
\usepackage{booktabs}
\usepackage{cleveref}
\usepackage{graphicx}

\newcommand{\lin}{\text{lin}}

\newcommand{\citep}{\cite}

\renewcommand{\paragraph}[1]{\vspace{.5em}\noindent\textbf{#1}}

\newcommand{\D}{\mathcal{D}}
\renewcommand{\test}{\text{test}}

\title{
LQF: Linear Quadratic Fine-Tuning \\ %

}

\author{Alessandro Achille$^1$ \ \ Aditya Golatkar$^{2,1}$ \ \ Avinash Ravichandran$^1$ \ \ Marzia Polito$^1$ \ \ Stefano Soatto$^1$\\
$^1$Amazon Web Services \ \ \ $^2$UCLA\\
\texttt{\{aachille,ravinash,mpolito,soattos\}@amazon.com aditya29@cs.ucla.edu}
}

\begin{document}

\maketitle

\begin{abstract}

Classifiers that are linear in their parameters, and trained by optimizing a convex loss function, have predictable behavior with respect to changes in the training data, initial conditions, and optimization. Such desirable properties are absent in deep neural networks (DNNs), typically trained by non-linear fine-tuning of a pre-trained model. Previous attempts to linearize DNNs have led to interesting theoretical insights, but have not impacted the practice due to the substantial performance gap compared to standard non-linear optimization. We present the first method for linearizing a pre-trained model that achieves comparable performance to non-linear fine-tuning on most of real-world image classification tasks tested, thus enjoying the interpretability of linear models without incurring punishing losses in performance. LQF consists of simple modifications to the architecture, loss function and optimization typically used for classification: Leaky-ReLU instead of ReLU, mean squared loss instead of cross-entropy, and pre-conditioning using Kronecker factorization. None of these changes in isolation is sufficient to approach the performance of non-linear fine-tuning. When used in combination, they allow us to reach comparable performance, and even superior in the low-data regime, while enjoying the simplicity, robustness and interpretability of linear-quadratic optimization.
\end{abstract}

\section{Introduction}

Deep neural networks (DNNs) are powerful but finicky. They can carve complex decision boundaries through high dimensional data such as images, but even small changes in the training set, regularization method, or choice of hyperparameters can lead to vastly different outcomes. This phenomenon, %
typical of highly non-linear optimization, is observed even when fine-tuning a pre-trained model, which is the most common {\em modus operandi} in practice: Starting from a DNN trained on some dataset, a few steps of stochastic gradient descent (SGD) are used to minimize a loss function computed on a another dataset. This is unlike models whose parameters are found via convex optimization, such as support-vector machines: They have a global optimum, found from any initial condition, and small changes in the data, the regularization scheme, and hyperparameters yield small and interpretable changes in the final solution.

\vspace{.5em}

Lack of robustness to training conditions, and opaqueness of the resulting model, appear to be the price to pay for more performing and expressive classifiers such as a DNNs. This price  is measured in time and cost of hyperparameter optimization (HPO). For example, simply changing the multiplier for weight decay requires retraining from scratch,
as optimizing the new loss starting from the previous solution gives suboptimal results \cite{golatkar2019time}. The complex relation between training data and final model makes it impossible to predict the effect of individual data, renders most generalization bounds vacuous, and makes it hard to impose even simple constraints, such as those arising from fairness criteria \cite{kearns2019ethical} or backward compatibility \cite{shen2020towards}.

\begin{figure}
    \centering
    \includegraphics[width=\linewidth]{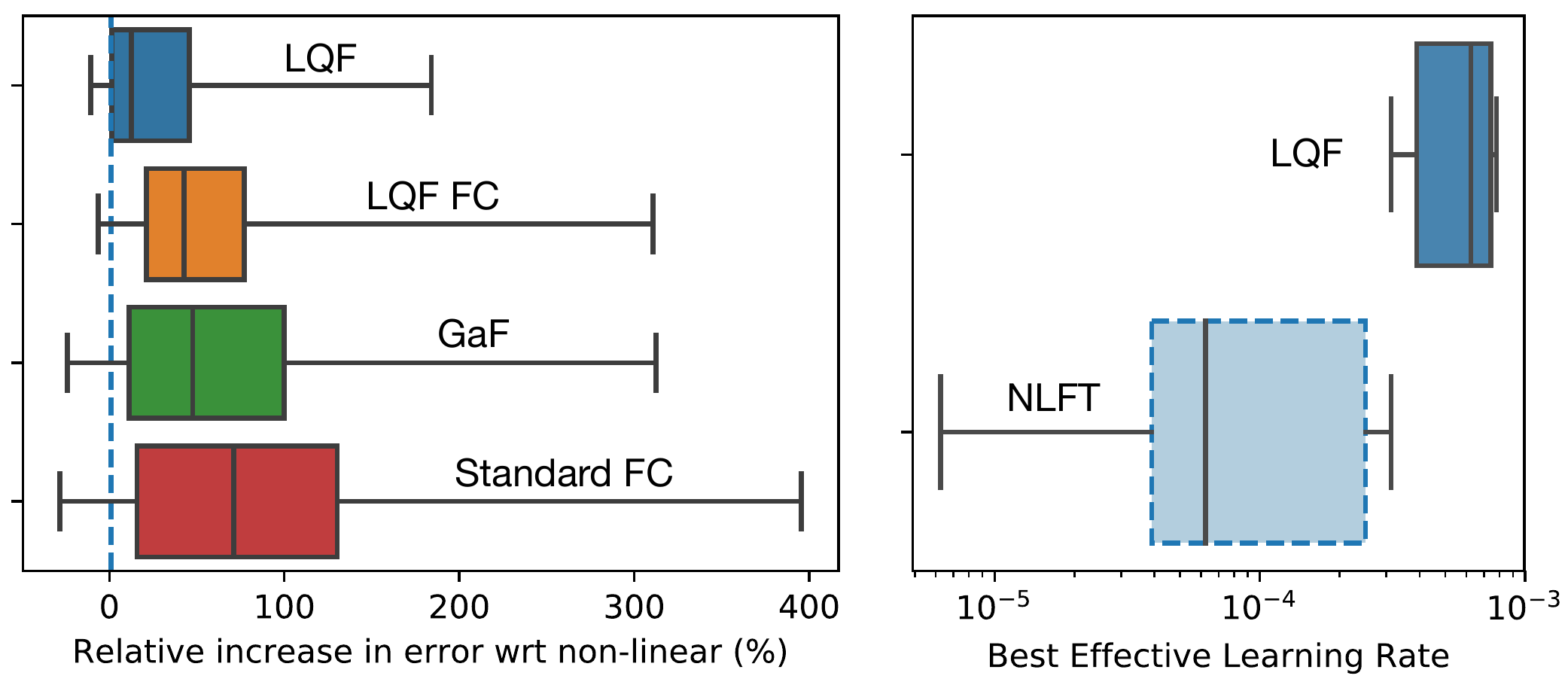}
    \caption{\textbf{Linear vs.\ nonlinear fine-tuning {\footnotesize{(NLFT).}}} (\textbf{Left}) Box-plot of the distribution of test errors achieved by different linearization methods on the datasets in \Cref{table:dataset-errors}, relative to the error achieved by NLFT (dashed line at origin). Whiskers show best/worst results, boxes extend from lower to upper quartiles (the results on half of the datasets are concentrated in the box), the central line represents the median increase in error.
    GaF \cite{Mu2020Gradients} (green, 47\% median error increase) is better than training a linear classifier on a fixed embedding (red, 71\% increase), but slightly worse than LQF applied just to that linear classifier (LQF FC, orange, 42\% increase). LQF is the closest linear method to the non-linear paragon (blue, 12\% increase).
    (\textbf{Right}) \linebreak We show the distribution of best effective learning rates as the task varies. While for NLFT we need to search in a wide range to find the optimal training  parameters for a task (wide dashed box), for LQF the same learning rate works almost equally well for all tasks (narrow solid box).
    }
    \label{fig:comparison}
\end{figure}

\vspace{.5em}

The desire to make their training more robust and interpretable has led some to {\em linearize} DNN models around an initial set of weights \cite{lee2019wide}. But while this has led to interesting theoretical insights, the analysis has failed to yield improvements in the practice. In particular, \Cref{fig:comparison} shows that linearized models perform large-scale image classification marginally better than simply training a linear classifier on a fixed pre-trained embedding. Non-linear fine-tuning with exhaustive HPO remains the performance paragon.
The trade-off between performance and robustness, %
typical of many complex systems, is a manifestation of the classic bias-variance tradeoff: By reducing the sensitivity of the trained model to perturbations in the parameters or training data (variance \Cref{fig:comparison}, right), we incur a decrease in average performance (bias, left). While we do not expect linearized models to outperform non-linear fine-tuned ones -- except in cases where the latter fails to optimize correctly (\Cref{sec:online}) -- in this paper we explore how far we can push their accuracy, so we can enjoy the robustness and interpretability of linear models without punishing performance loss.

\paragraph{Our contribution}, {\em Linear Quadratic Fine-Tuning} (LQF), is a method to linearize and train deep neural networks that achieves
comparable performance to non-linear fine-tuning (NLFT) on real-world image classification, while enjoying all the benefits of linear-quadratic optimization. LQF performs fine-tuning without optimizing hyper-parameters such as learning rate or batch size, enables predicting the effect of even individual training samples on the trained classifier, and easily allows incorporating linear constraints during training. %
LQF achieves performance comparable to NLFT, and better in the low-data regime which is the most relevant to real applications.

The key enablers of LQF are simple and known in the literature, although not frequently used: (i) We replace the cross-entropy loss with the mean-squared error loss, making the optimization problem quadratic \cite{golik2013cross,hui2020evaluation,barz2020deep}, (ii) we replace ReLU with Leaky-ReLU \cite{maas2013rectifier}, and (iii) we perform pre-conditioning using Kronecker factorization (K-FAC) \cite{martens2015optimizing}. Individually, these changes bring limited improvements to standard training of non-linear models. However, we show that their combined use has a much larger impact on the performance of linearized models (\Cref{fig:ablation}). %

\section{Related Work}

\noindent\textbf{Network linearization} uses a standard first-order Taylor expansion of a pre-trained model (eq.~\ref{eq:taylor}).
Work on the Neural Tangent Kernel (NTK) showed that linear models approximate the dynamics of randomly initialized deep networks \cite{jacot2018neural,lee2019wide}, at least when the number of filters goes to infinity, although   \cite{Mu2020Gradients} argue the analysis is valid for finite networks in the case of fine-tuning when the weights $w$ are close to those of the pre-trained model $w_0$. Using the linearized model amounts to training a linear model using the gradients of the original network as features, which is challenging given their dimension. However, \cite{Mu2020Gradients} shows that this can be done efficiently with a modified forward pass using the Jacobian-Vector product algorithm. But while the training cost of the linear model is comparable to that of the original one, performance is not (\Cref{fig:comparison}). \cite{bai2020taylorized} extends the approximation to higher orders and show increased fidelity, but loses the linearity of the model.

\paragraph{Loss function.} We use the mean squared error (MSE) for fine-tuning instead of the more common cross-entropy (CE) loss, even when pre-training was performed with the latter.
This may seem counter-intuitive, but there is growing evidence that the MSE can be competitive to train classifiers \cite{golik2013cross,barz2020deep,hui2020evaluation} and is just as well grounded theoretically \cite{csiszar1991least}. There is also evidence that the standard cross-entropy loss learns representations that are more transferable \cite{kornblith2020whats}, and hence it is better suited for pre-training, even if it is not necessarily the best performing loss in terms of test accuracy.
Our use of CE for pretraining and MSE for fine-tuning fits in this paradigm.

\paragraph{Pre-conditioning} is important to converge on badly conditioned convex problem. We use Kronecker-factorized (K-FAC) approximation of the Fisher \cite{martens2015optimizing,george2018fast,grosse2016kronecker} to efficiently approximate the curvature of the loss function. Since curvature is constant in LQF, we only need to compute the preconditioning matrix once, sensibly reducing its cost.

\paragraph{Hyper-parameters} can significantly affect performance in NLFT, which requires careful optimization (HPO) of learning rate, momentum and batch size based on the source and target task \cite{li2020rethinking}. A linear model does not require extensive HPO  (\Cref{fig:comparison}). Being strongly convex, LQF has a unique global minimum to which SGD will provably converge, assuming a proper learning rate schedule is used \cite{nesterov2013gradient}.

\paragraph{Interpretability.} We study how a given training sample affects the learned model and its predictions. This analysis is similar to that of \textit{influence functions} \cite{koh2017understanding,basu2020influence}, but in our case can be done without approximation since our loss is quadratic. Unlike influence functions, we can also efficiently compute the change of activations on a validation set, which we use to compute informativeness (\Cref{sec:intepretability}).

\paragraph{Applications.} Having a linear model that performs on-par with NLFT makes it possible to tackle open issues in large-scale DNNs, such as
(i) backward-compatibile training without model distillation \cite{shen2020towards}, (ii)  leave-one-out test error estimation without a test set \cite{giordano2019swiss} (iii)    measuring the influence of individual samples on the trained model \cite{koh2017understanding,anonymous2021estimating} to enable active learning and enforce privacy guarantees \cite{dwork2008differential}, (iv) integrating the model in a continual learning framework (\Cref{sec:online}). %

\section{Linear-Quadratic Fine-tuning}
\label{sec:method}

\paragraph{Linearized model.} LQF addresses the lack of robustness and interpretability of non-linear deep models by replacing the network itself with  a first-order linear approximation.
Let $f_w(x)$ denote the output of a deep network with weighs $w$ on an input image $x$, let $w_0$ denotes an initial set of weights, for example obtained after pre-training on ImageNet. We consider the \textit{linearization} $f^\lin_{w}(x)$ of the network $f_w(x)$ given by the first-order Taylor expansion of $f_w(x)$ around $w_0$:
\begin{align}
\label{eq:taylor}
    f^\lin_{w}(x) = f_{w_0}(x) + \nabla_w f_{w_0}(x) \cdot (w - w_0).
\end{align}
Note that $f^\lin_{w}(x)$ is linear with respect to the weights $w$; however, due to the non-linear activation functions, it is still a highly non-linear function of the input $x$.
If the approximation is accurate, ideally we want the linearized $f^\lin_w$ and the original $f_w$ to reach similar accuracy when trained:
\begin{equation}
\label{eq:approx}
    \operatorname{accuracy}(f^\lin_{\hat{w}}, \D_\text{test}) \approx \operatorname{accuracy}(f_w, \D_\text{test})
\end{equation}
where $\hat{w}$ and $w$ are obtained by minimizing, respectively, the loss of the linear and non-linear model on a training set $\D$.
The obvious advantage of the linear model is that, for strongly convex loss functions, the global optimum $\bar w$ is unique and, for a quadratic loss functions, can even be written in closed form.
So, if \cref{eq:approx} was satisfied, we would stand to gain in terms of speed of fine-tuning, performance (convergence is guaranteed to the global optimum), and interpretability (the effect of a training sample on the test prediction can be computed exactly).

\paragraph{Training a linearized model} requires computing
the product of the Jacobian $\nabla_w f_{w_0}(x)$ with the weights $w$, which would require a separate backward pass for each sample.
However, \cite{Mu2020Gradients} shows that using Gradients as Features (GaF) the Jacobian-Vector product of \cref{eq:taylor} can be computed with a modified forward pass, at a cost comparable to that of the running the original network.

\paragraph{Accuracy of linearized models.}
In \Cref{fig:comparison} and \Cref{table:dataset-errors} we show that the accuracy of linearized models
suffers compared to the paragon of NLFT. In most datasets, even fine-tuning just the linear classifier (fully-connected, or FC, layer) is only slightly worse than GaF, thus violating \eqref{eq:approx}. We now analyze the differences that led to these differences between the dynamics of fine-tuning a non-linear network and its linearized version, and propose changes that lead to LQF.

\begin{figure}
    \centering
    \includegraphics[width=.8\linewidth]{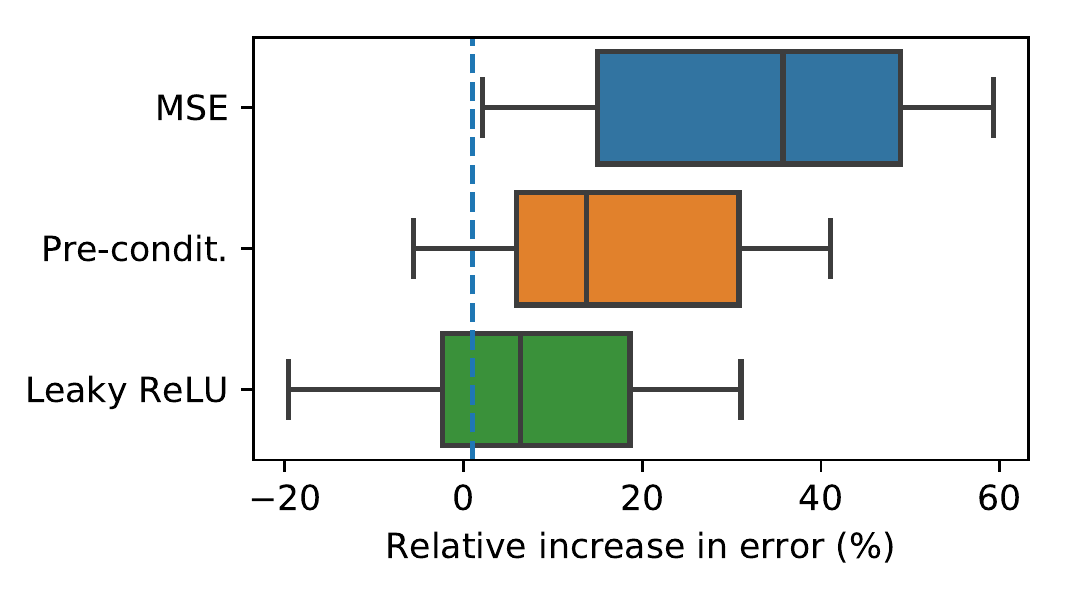}
    \caption{\textbf{Ablation study.} Box plot showing relative error increase on the datasets in \Cref{table:dataset-errors} when removing consititutive components of LQF. Removing MSE in favor of the standard CE loss yields the largest increase in error (blue). Removing K-FAC pre-conditioning, leaving standard SGD (orange), has also a sizeable effect.
   Finally, on most datasets, Leaky-ReLU performs better than standard ReLU (green).
    }
    \label{fig:ablation}
\end{figure}

\subsection{Loss function}
\label{sec:mse-loss}

We use a regularized mean-squared error (MSE) loss, even if the original model is pre-trained by minimizing the regularized cross-entropy loss:
\begin{equation}
\label{eq:mse-loss}
    L_{\rm MSE}(f; {\cal D}) =  \sum_{(x_i, y_i) \in {\cal D}}  \| \alpha y_i - f(x_i)\|^2 +\frac{\lambda}{2} \|w - w_0\|^2,
\end{equation}
where $y_i$ is the one-hot vector encoding the class label and $\alpha$  is a scaling factor for the targets  (we set $\alpha=15$ as in \cite{hui2020evaluation}).
There are several reasons to make this change. First, while standard DNNs have several normalization layers (such as batch normalization) that keep the output bounded, the linear model $f^\lin_w(x)$ in \cref{eq:taylor} directly outputs the dot product $\nabla_w f_{w_0}(x) \cdot w$, which involves tens of millions of weights. This product can easily grow  large and saturate the softmax of the cross-entropy loss, making the training process unstable. When that happens, we observe that only the last layer features are effectively trained.
On the other hand, the MSE loss does not saturate and forces the output to remain close to the target one-hot vectors, and hence bounded. We also note that using MSE  gives better results \textit{even} for standard non-linear training (see Appendix), especially when training on small datasets.%

A secondary effect of using the MSE loss in conjunction with a linear model is that we can write the unique global optimum in closed form:
\begin{align}
\label{eq:mse-optimum}
w^* = (J^T J + \lambda I)^{-1} J (Y - f_0(X))
\end{align}
where $J \equiv \nabla_w f_{w_0}(X)^T$ is the matrix of the Jacobians of all the samples in the training set and $Y$ is the matrix of all target vectors. While $w^*$ cannot be computed directly using this formula due to large size of the matrices, in \Cref{sec:intepretability} we show that \cref{eq:mse-optimum} allows us to easily compute the influence of a given training sample on the test predictions, enhancing interpretability of the training process.

\subsection{Pre-conditioning with K-FAC}
\label{sec:kfac}

The MSE loss in \cref{eq:mse-loss} is strictly convex, so SGD is guaranteed to converge to the unique global minimum, provided an appropriate learning rate decaying schedule is used \cite{nesterov2013gradient}. However, %
different directions in the loss landscape  may have different curvature, slowing convergence: the same learning rate may be too fast high-curvature directions and too small for flat ones. Convergence speed is governed by the condition number of the Hessian matrix (ratio between the largest and smallest eigenvalues, i.e., maximum and minimum curvature), which can be modified by multiplying the update with a {\em pre-conditioning} matrix $A_t$
\begin{align}
    w_{t+1} = w_t - \eta A_t g_t,
\end{align}
where $g_t$ is the batch gradient computed at step $t$ and $\eta$ is the learning rate.
The matrix $A_t$ allows training at different speed in each direction.
In a quadratic problem with constant preconditioning $A_t = A$, the expected distance from the optimum $w^*$ at time $t$ using adaptive SGD is given by:
\begin{align}
    \E[w_t - w^*] = (I - \eta A H)^t (w_0 - w^*),
    \label{eq:adaptive-sgd-time-t}
\end{align}
where $H$ is the (constant) Hessian of the loss function. In particular, the fastest convergence is obtained when $A = H^{-1}$ is the inverse of the hessian. For this reason, most adaptive methods try to approximate $H^{-1}$.

\begin{figure}
    \centering
    \includegraphics[width=.8\linewidth]{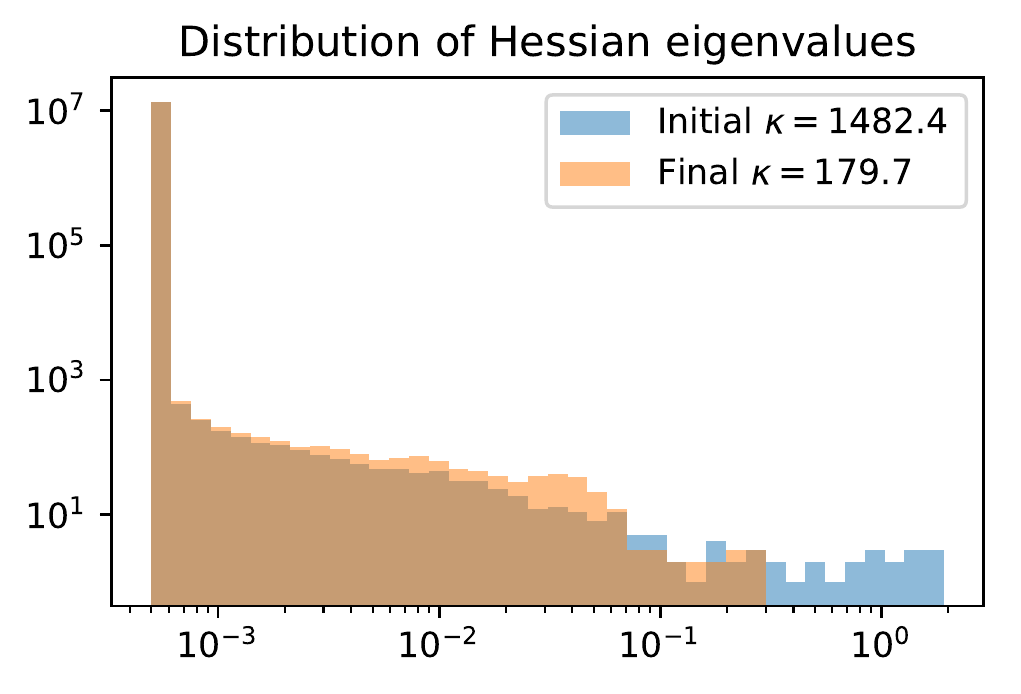}
    \caption{\textbf{Unlike SGD, LQF requires pre-conditioning.} Histogram of the eigenvalues of the Hessian loss function at initialization (blue) and at the end of optimization of a non-linear network (orange). Note that at initialization a few directions have very large eigenvalues while many directions have much smaller eigenvalues, resulting in a high condition number $\kappa = 1482.4$. However, SGD naturally moves to areas of the loss landscape that are better conditioned, with $\kappa = 179.7$ at the end of training (note that the largest eigenvalues decrease), thus making convergence easier. This ``automatic conditioning'' makes pre-conditioning un-necessary in non-linear networks. On the other hand, since in LQF the Hessian is fixed throughout the training, we need to use K-FAC pre-conditioning to facilitate convergence.
    }
    \label{fig:eigenvalues}
\end{figure}

Adaptive methods are commonplace in convex optimization and some \cite{kingma2014adam} are widely used in deep learning. However, for large-scale visual classification, standard SGD with momentum performs comparably or better than adaptive methods. We hypothesize that this may be due to the changing curvature during training, which causes the condition number to decrease, making the problem well-conditioned even without adaptation (see \Cref{fig:eigenvalues}). However, this cannot happen when training a linearized network, since the curvature of \cref{eq:mse-loss} is constant and fully determined in pre-training. For this reason, we add a pre-conditioner for \cref{eq:mse-loss} in LQF.

Note that, in our case, the Hessian coincides with the Fisher Information Matrix (FIM). We use the Kronecker-Factorized (K-FAC) approximation of the FIM as pre-conditioner \cite{martens2015optimizing}. Unlike diagonal approximations, K-FAC allows us to estimate off-diagonal while remaining tractable.

\paragraph{K-FAC is efficient for LQF.} K-FAC may make the weights converge too quickly to suboptimal sharp minima that do not generalize well and, since the curvature changes over time, the approximation of the Hessian needs to be updated. However, in LQF there is a unique minimizer and the curvature is constant. This makes the use of K-FAC especially efficient in our setting. Compared to other adaptive methods like Adam \cite{kingma2014adam}, in our setting K-FAC provides an accurate estimation of the Hessian inverse which can be used to interpret the effect of a training sample (\Cref{sec:intepretability}) and to easy select an appropriate learning rate.

\paragraph{Learning rate selection} requires the norm of the eigenvalues of $I - \eta AH^{-1}$ to be smaller than one. If $A$ is a good approximation of $H$, the eigenvalues of $AH^{-1}$ are centered around $1$ and $\eta=1$ is optimal. Since we need $\eta < 2/\lambda_\text{max}(A^{-1}H)$ to guarantee convergence -- and we expect $\lambda_\text{max}(A^{-1}H) \approx 1$ --  we choose a conservative $\eta = 0.1$, which gives consistent performance on all datasets without further tuning (\Cref{fig:comparison}, right).

\subsection{Leaky-ReLU activations}

Rectified linear units (ReLUs) \cite{nair2010rectified,sun2015deeply}
leave the positive component of the input unchanged and replace the negative component with zero, which blocks information from certain weights. This is not harmful during normal training, since batch normalization helps avoiding dead filters. However, in  linearized models only the gradient of the filter at initialization matters in computing the Jacobian-Vector Product. If this is zero, it will stunt training for that filter.
To prevent this, we simply replace ReLU with Leaky-ReLu \cite{maas2013rectifier}, which does not annihilate the negative component but simply replaces it with a value proportional to the input. %
Another advantage of Leaky-ReLU is that its first order approximation near zero is more accurate than that of ReLU, making the first-order approximation in \cref{eq:taylor} more accurate.

\paragraph{Performance of LQF.}
After accounting for these changes, we repeat the previous experiments on different datasets and report the overall results in
\Cref{fig:comparison} and detailed  in \Cref{table:dataset-errors}. LQF outperforms other linearization methods on all the datasets, and achieves performance comparable to non-linear fine-tuning on most. In the low data regime, the strong inductive bias of the linear model acts as a regularizer and the linearized model outperforms the non-linear fine-tuned one (\Cref{fig:low-shot}). We discuss the results in detail in \Cref{sec:experiments}.

\subsection{Interpretability}
\label{sec:intepretability}

LQF gives us an efficient way of estimating the influence, or ``effect'' of a given training sample on the final weights and on the test predictions of the DNN, as well as measuring the informativeness to the training process. This can be used for dataset summarization  (Sect.~\ref{sec:summarization}) and to interpret model predictions: Had a particular sample (cause) not been used for training, how would the final model (effect) differ? Using  \cref{eq:mse-optimum} we have the following expression for the weights $w_{-i}$ we would have obtained if training without the sample $x_i$:
\begin{align}
\label{eq:weight-change}
    w_{-i}^* = w^* + (F_{-i} + \lambda I)^{-1} g_i.
\end{align}
where $g_i \equiv \nabla_{w^*} f_w(x)$ is the Jacobian of the $i$-th sample, $e_i \equiv \frac{1}{N} (y - w^* g_i)$ is the weighted prediction error on the sample $x_i$, $N$ is the number of training samples, and $F_{-i}$ is the hessian computed without $x_i$.

Using \cref{eq:weight-change} we can estimate how much the prediction $f_w(x_\test)$ on a test sample $x_\test$ would change had we not seen the training sample $x_i$. That is, we can measure which training sample has contributed more to a correct or incorrect decision of the network. Using \cref{eq:weight-change}, in the Appendix we show that
\begin{align}
    &f_{w^*}(x_\test) - f_{w^*_{-i}}(x_\test) \approx \nonumber\\
&\hspace{3em} (1 - \frac{\alpha}{N-1 + \alpha}) e_i g_\test^T (F + \lambda I)^{-1} g_i.
  \label{eq:activations-change}
\end{align}
Since K-FAC already provides an approximation of $(F + \lambda I)^{-1}$, we can estimate the expression efficiently.

More generally, one can define the unique functional information \cite{anonymous2021estimating} of a training sample $x_i$ as the expected influence that the sample has on the network output on a validation set:
\begin{align}
\operatorname{F-SI}((x_i, y_i)) = \E_{x \sim \D_\text{val}}\big[\|f_w(x) - f_{w_{-i}}(x)\|^2\big]
\label{eq:functional-information}
\end{align}
This measure -- which can be used for dataset summarization and to interpret what the network learn from a training samples (\Cref{sec:summarization}) -- is also easily estimated  using \cref{eq:activations-change}.

\section{Results}
\label{sec:experiments}

We test LQF on several standard vision classification tasks used to benchmark fine-tuning learning \cite{li2018explicit,li2020rethinking} and additional tasks to test cross-domain transfer (see Appendix). We find that LQF outperforms other linearization methods on almost all tasks. LQF also outperforms standard non-linear training (NLFT) in tasks such as few-shot learning and on-line learning, where scarce data causes training instabilities. We also illustrate the benefits of interpretability.

\paragraph{Architecture and training.} As the base model for linearization we use a ResNet-50 with Leaky-ReLU pre-trained on ImageNet. We implement the forward pass of \cref{eq:taylor} using  \cite{Mu2020Gradients}, and minimize the MSE loss, with K-FAC pre-conditioning (using the implementation of \cite{george2018fast}). For NLFT, we also use a pre-trained ResNet-50 with ReLU activations, and minimize the cross-entropy loss. All experiments are conducted with SGD with  momentum 0.9, learning rates $\eta \in \{0.01,0.001\}$, weight decay $\in \{10^{-4}, 10^{-5}\}$ and report the best results. We discuss further training details in the Appendix.

\subsection{Comparison}
\label{sec:comparison}

Ideally, LQF should achieve close performance to NLFT on all tasks. We also compare with two other linearized a deep network: Standard FC, which uses the pre-trained network as a feature and only retrains the last FC layer, and Gradients as Features (GaF) \cite{Mu2020Gradients} which linearizes the network without any of the changes described in \Cref{sec:method}. We  also introduce a further baseline, LQF FC, which uses the changes in \Cref{sec:method}, but only to train the final FC layer.

We train each model on several ``coarse-grained'' datasets, covering different tasks and domains and report the results in \Cref{table:dataset-errors}. In \Cref{fig:comparison} we show the distribution of error of each methods relative to the NLFT paragon. We see that GaF improves only marginally over the  FC baseline, and is usually far from the non-linear performance. On the other hand, LQF performs comparably with NLFT on all the coarse-grained datasets. Surprisingly, training only the final classifier using LQF (LQF FC) improves over Standard FC, suggesting that the changes discussed in \Cref{sec:method} are helpful in broader scenarios.

\begin{table}
    \centering
    \small
\setlength{\tabcolsep}{4pt}
\begin{tabular}{l|c|cccc}
\toprule
 &  NLFT &    LQF$^*$  &  LQF FC$^*$ &    GaF &  FC \\
\hline
\hline
Caltech-256 \cite{griffin2019caltech} &      14.1 & \textbf{14.5} &    17.4 & 15.4 &         15.7 \\
Chest X-Ray  \cite{kermany2018identifying}       &       4.6 &  \textbf{7.1} &     8.3 &  9.5&         11.2%
\\
Malaria Cells \cite{rajaraman2018pre} &       3.2 &  \textbf{4.1} &     5.5 &  4.8 &          6.4%
\\
MIT-67  \cite{quattoni2009recognizing}           &      20.5 & \textbf{20.7} &    24.7 & 23.1 &         24.3 \\
Oxford Pets  \cite{parkhi12a}             &       6.7 &  \textbf{6.9} &     7.7 &  7.4 &          7.7 \\
\bottomrule
 \multicolumn{6}{c}{Fine-grained datasets (sorted by ImageNet distance)} \\
\hline
\hline
Stanf.\ Dogs \cite{KhoslaYaoJayadevaprakashFeiFei_FGVC2011}\!\!      &      13.8 & 12.4 &    12.9 & 10.5 &          \textbf{9.9} \\
Ox.\ Flowers  \cite{Nilsback06}\!\!  &       7.1 &  \textbf{7.1} &     9.7 & 13.3 &         13.6 \\
CUB-200 \cite{WelinderEtal2010}           &      19.6 & \textbf{24.0} &    29.1 & 28.4 &         29.2 \\
Aircrafts \cite{maji13fine-grained}   &      14.5 & \textbf{34.5} &    43.8 & 48.4 &         54.2 \\
Stanf.\ Cars \cite{KrauseStarkDengFei-Fei_3DRR2013}     &       9.6 & \textbf{27.1} &    39.2 & 39.4 &         47.3 \\
\bottomrule
\end{tabular}
    \caption{\textbf{Test error of linear and non-linear fine-tuning on coarse-and fine-grained classification}.  On almost all datasets LQF outperforms all other linear methods. Moreover, on most coarse datasets LQF is within 0.5\% absolute error from standard non-linear fine-tuning. Linear methods performs  worse than NLFT on fine-grained machine classification tasks that have a large domain gap with respect to ImageNet pretraining -- e.g., \textit{Stanford Cars} and \textit{FGVC Aircrafts}. However, even in this case LQF reduces the error by up to 20\% with respect to competing methods.
    }
    \label{table:dataset-errors}
\end{table}

\subsection{Fine-grained classification and Bilinear Pooling}
\label{sec:fine-grained}

Since LQF is based on a first-order Taylor expansion, it has an implicit bias toward remaining close to the pre-training (ImageNet in our case).
\cite{li2020rethinking} notes that such a bias may improves accuracy on tasks that are easier or nearby (e.g., Stanford Dogs, Caltech-256), but it is detrimental on fine-grained tasks with a large domain gap from ImageNet (e.g., CUB-200, Aircrafts, Cars). We expect the same to hold LQF. Indeed, in \Cref{table:dataset-errors} we show that while performance is close or better on similar tasks, it degrades for farther tasks (sorted as in \cite{li2020rethinking}).  However, even in this case, we observe that LQF significantly outperforms other linearization techniques.

\paragraph{B-LQF.}
These results  suggest that finding a pre-training closer to the target task may help improve the linearizaion performance \cite{achille2019task2vec,nguyen2020leep,cui2018large}, which is indeed an intereting direction of research. On the other hand,
\cite{lin2015bilinear} suggests that even for generic ImageNet pretraining, the covariance of the last-layer features, rather than their mean, is more suited as a feature for fine-grained visual classification. Following this intuition, we replace the last layer global spatial mean pooling with square root bilinear pooling: if $z_{i,j}$ are the features of the last-layer feature map, instead of feeding their mean $\mu = \E_{i,j}[z_{i,j}]$ to the fully connected layer, we feed the square root $\sqrt{\Sigma}$ of their covariance matrix $\Sigma$ \cite{lin2015bilinear,lin2017improved,george2018fast}. In \Cref{tab:fine-grained} we see that -- while bilinear pooling  does not improve the performace of NLFT too much -- combining bilinear-pooling with LQF (B-LQF) significantly boosts its performance on fine-grained tasks.

\subsection{Ablation study}

We measure the effect of each component of LQF on its final accuracy. In \Cref{fig:ablation} we show the relative increase in error when we ablate the changes discussed in \Cref{sec:method}. Using the MSE loss instead of cross-entropy yields the largest improvement. Using K-FAC also carries significant weight in the final solution. This is interesting since, in principle, SGD without pre-conditioning should recover the same minimum. This reinforces the point that the optimization problem is badly conditioned (\Cref{fig:eigenvalues}), and  proper pre-conditioning is necessary. Using Leaky-ReLU activations instead of ReLUs also improve the final results on average, even if by a lesser margin than the other two changes.

\begin{table}
    \centering
    \small
    \setlength\tabcolsep{4.8pt}
\begin{tabular}{l|cc|ccc}
\toprule
 &  NLFT & MPN-COV &    B-LQF$^*$  &  LQF$^*$ & GaF\\
\hline
\hline
CUB-200            &    19.6 & 17.0 &  \textbf{17.1} & 24.0 &  29.2\\
Aircrafts        &   14.5  & 15.4 & \textbf{29.2} & 34.5 & 54.2\\
Cars           &     9.6 & 9.1 &   \textbf{16.3} &  27.1 & 47.3 \\
\bottomrule
\end{tabular}
    \caption{\textbf{Fine-grained classification and bilinear pooling.} Normalized bilinear pooling (MPN-COV) \cite{lin2017improved,george2018fast}  does not significantly improve ordinary classification results (NLFT) in our setting. However, LQF with bilinear pooling (B-LQF) significantly improves test  accuracy compared to average pooling (LQF).
    }
    \label{tab:fine-grained}
\end{table}
\subsection{Low-shot data regime}

\begin{figure}
    \centering
    \includegraphics[width=.95\linewidth]{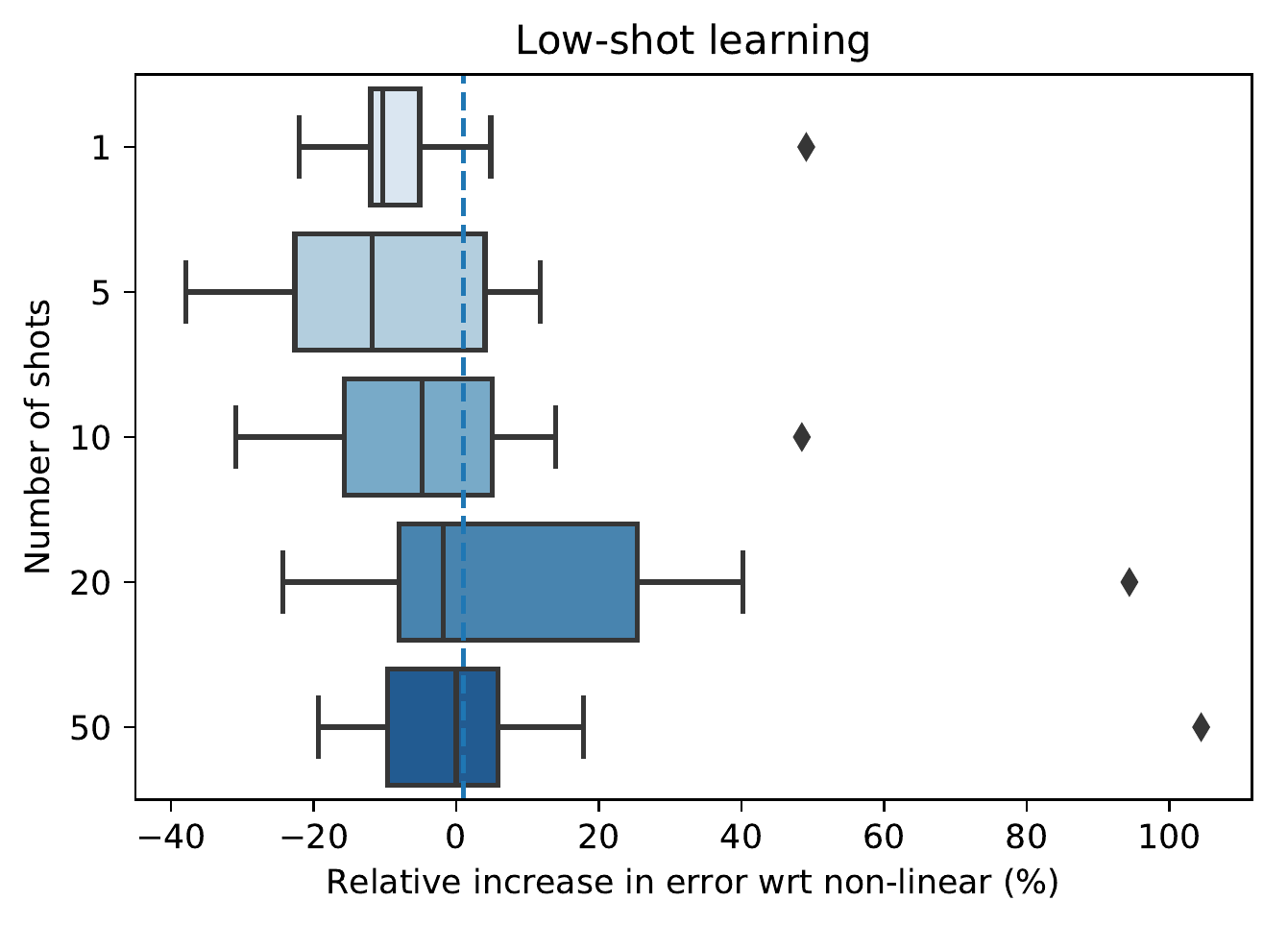}
    \caption{\textbf{Efficiency of LQF at different data regimes.} Comparison of LQF and standard fine-tuning for different number of training samples per category (``shots''). On average, LQF performs better than standard fine-tuning when the datasets are relatively small. The single outlier is FGVC Aircraft, where NLFT tends to perform better (see \Cref{sec:fine-grained}).
    }
    \label{fig:low-shot}
\end{figure}

LQF, like any linearization, cannot be expected to outperform NLFT in general. However, there may be scenarios where it does systematically. In the low-data regime, non-linear networks can easily memorize samples using spurious correlation without having a chance to converge to fit relevant features.  On the other hand, LQF should have a strong inductive bias arising from the existence of a unique global minimum. To test this, in \Cref{fig:low-shot} we plot the test error of LQF relative to the NLFT paragon. In the $k$-shot scenario (i.e., training using only $k$ samples per class) LQF outperforms NLFT systematically. This suggests that LQF may be a better base model than NLFT in few-shot learning, also because it facilitates the incorporation of any convex regularizer.

\subsection{On-line learning}
\label{sec:online}

In many applications, training data is not all available at the outset. Instead, it trickles in continuously and triggers re-training upon reaching a sufficiently large volume. This protocol engenders tradeoffs between cost and obsolescence. Fine-tuning with relatively small batches is a typical compromise, but can lead to suboptimal solutions when the network  memorizes the new samples or catastrophically forgets old ones. LQF is not subject to this tradeoff as it has a unique minimum that is adjusted incrementally as new data is received.

\begin{figure}[b]
    \centering
    \includegraphics[width=.85\linewidth]{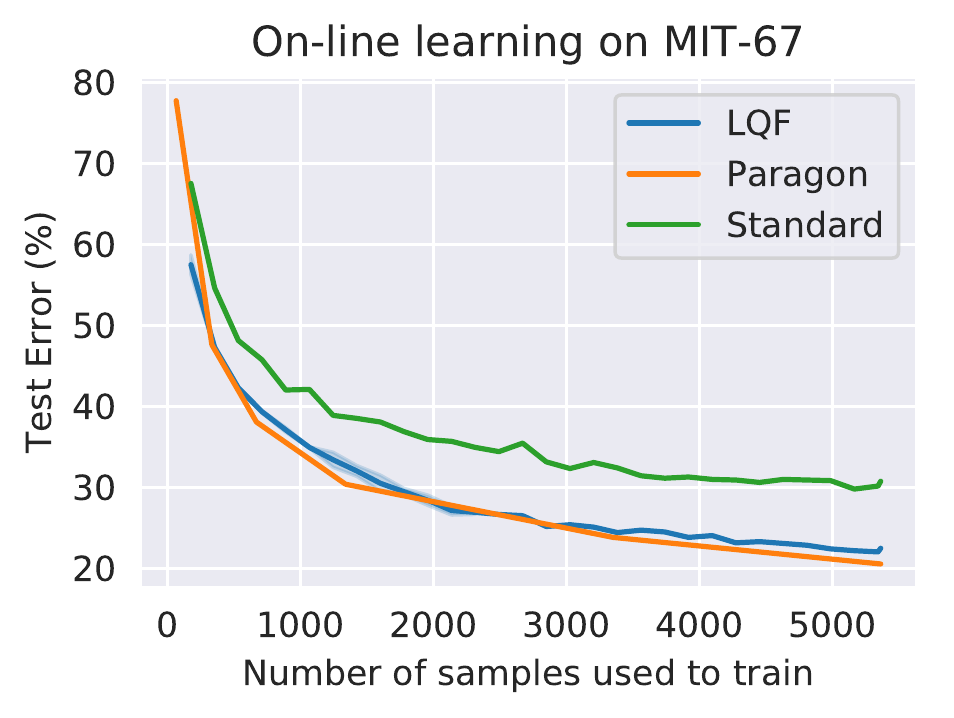}
    \caption{\textbf{Comparison of on-line learning methods.} Standard non-liner networks may converge to a bad local minimum when trained incrementally, and consequently have a sub-optimal performance compared to a non-linear network re-trained from scratch every time more data is obtained (paragon). On the other hand, LQF has a unique minimum to which it is guaranteed to converge, ensuring that the performance will be close to the paragon at all steps.
    }
    \label{fig:online-learning}
\end{figure}

In \Cref{fig:online-learning}, we compare incremental LQF incremental re-training with the NLFT paragon of re-training from scratch every time we obtainmore data on the MIT-67 dataset. Let $\D_t$ be the dataset containing all the samples seen up to step $t$; we train on $\D_t$ starting from the weights $w_{t-1}$ obtained at the previous step and stop training when the train error on $\D_t$ is below 0.5\% for 5 consecutive epochs. In \Cref{fig:online-learning}, we show the test accuracy of the model at different  steps. As expected, incremental NLFT performs substantially worse than the paragon, as it gets stuck in suboptimal local minima. On the other hand, LQF is always close to the paragon even if trained incrementally while being much cheaper to train since we do not need to retrain from scratch.

\subsection{Informativeness of samples}
\label{sec:summarization}

\begin{figure*}
    \centering
    \includegraphics[trim=55 0 0 0, clip, width=.63\linewidth]{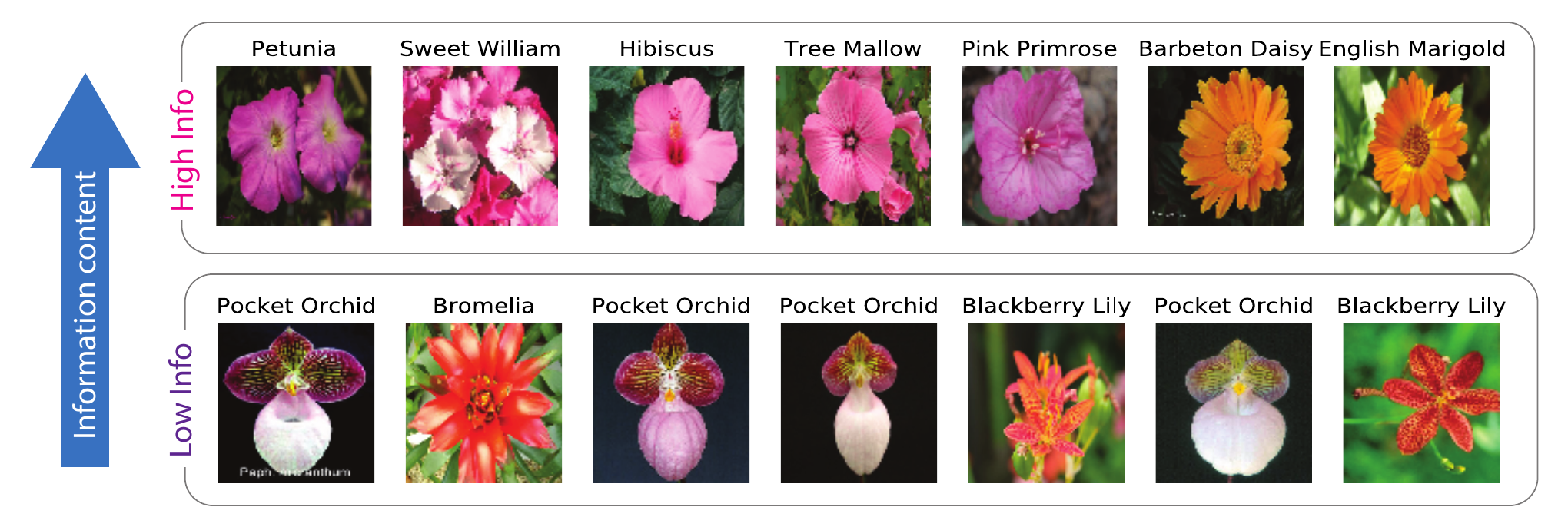}%
    \hspace{1em}%
    \includegraphics[width=.33\linewidth]{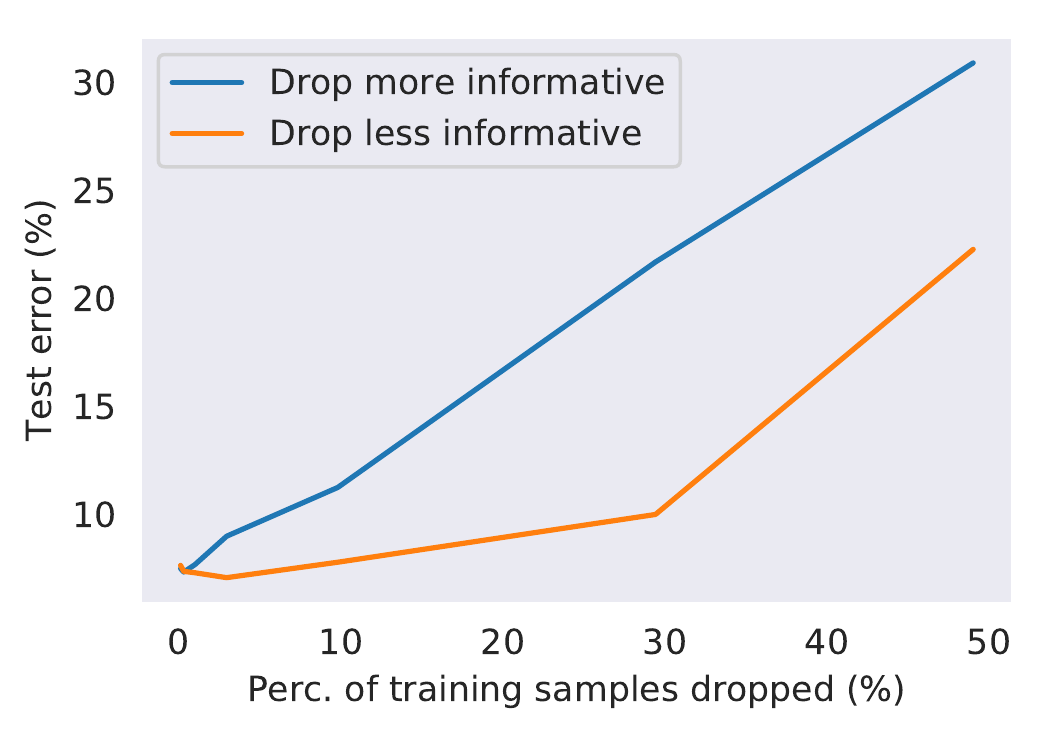}
    \caption{\textbf{Informativeness of samples.} \textbf{(Top  row)} Seven representative  examples from the 25 most informative images for a network trained on Oxford Flowers (complete set in appendix). The network considers more informative samples of flowers that are hard to distinguish, such as \textit{English Marigold} and \textit{Barbeton Daisy}, or \textit{Hibiscus}, \textit{Petunia} and \textit{Pink Primrose}. On the other hand, images of flowers that have a distinctive shape (e.g., \textit{Orchid}) or near duplicated images are not informative for the training \textbf{(bottom row)}.
    \textbf{(Right)} \textbf{Dataset summarization.} We plot the final test accuracy when training LQF after dropping the $N$ most informative training examples (blue line) and the $N$ less informative (orange line). The test performance decreases faster when dropping informative examples, as we would expect, validating our informativeness measure.
    }
    \label{fig:information}
\end{figure*}

LQF is interpretable in that it allows measuring the effect of individual  data on the trained model, which can be used to summarize a dataset. We use \cref{eq:functional-information} to measure how informative a training sample is for the predictions of the final model. In \Cref{fig:information}, we display the most and least informative samples: Qualitatively, we observe that easy samples and near-duplicates are not informative. On the other hand, samples of classes that are easy to confuse are considered very informative by the model, since they affect the decision boundary. To test this, in \Cref{fig:information} (right) we plot the test error of the model when the most/least informative training samples are removed. As we expect, removing informative samples degrades the performance more than removing uninformative samples.

\subsection{Robustness to optimization hyper-parameters}
\label{sec:robustness}

Non-linear fine-tuning is highly sensitive to the choice of hyperparameters (\Cref{fig:comparison}, right). This has a direct impact on cost, as training requires broad hyper-parameter search. It would be desirable if one could fix hyperparameters once for fine-tuning. In \Cref{sec:kfac} we argued that, due to the use of MSE loss and K-FAC pre-conditioning, the optimal learning rate is independent of the task for LQF. When using SGD with momentum, the situation is slightly more complex, as we also need to select batch size $b$ and the momentum $m$. As already noted by \cite{li2020rethinking}, when fine-tuning, different runs typically have the same test accuracy as long as the effective learning rate $\text{ELR} = \frac{\eta}{(1 - m) b}$ is the same. We claim that, for LQF, there is a single ELR which is optimal for most tasks. To test this, we search the best learning rate $\eta \in \{0.05,0.01,0.005,0.001,0.0001\}$ and batch size $b \in \{16, 32, 64\}$, and compute the ELR of the best combination for each task. In \Cref{fig:comparison} (right) we report the distribution of effective learning rates that yield the best performance. The interval of  optimal ELRs for LQF is small, so one choice of ERL is nearly optimal for all tasks. We do not observe significant changes in test accuracy for minor variations of ELR. On the other hand,  for NLFT the ELR can vary by orders of magnitude depending on the task, and test performance can vary widely within that interval. Fine-grained search of the ELR is necessary for NLFT, and not LQF.  This robustness comes at the cost of a slight decrease in performance (\Cref{sec:comparison}, left), as expected due to the linearization.

\section{Discussion}

Usage of deep neural networks in practice often involves fine-tuning pre-trained models followed by heavy hyper-parameter optimization (HPO). This method can achieve good performance, but even minor variations of hyper-parameters can spoil performance. The low-data regime is also treacherous, as models can easily memorize the samples provided. It is difficult for non-expert to predict the effect of regularizers and hyper-parameters beyond trying it and seeing. Linear models, on the other hand, are supported by decades of theory that allows to predict the effect of changes in the data, or in the parameters, and bounds on the generalization error. However, their performance has not been comparable to that of black-box DNNs. LQF is a linear model that gets closer, to the point of making linearization viable beyond a theoretical analysis tool, as a practical method. The small drop in performance is offset by improved ease of training, interpretability, and cheaper hyper-parameter search.

LQF is not the solution to all fine-tuning problems. Being a first-order approximation, it can realistically be expected to work well when the task on which we wish to fine-tune is sufficiently close to the task on which the model was pre-trained. In practice, this limitation may be circumvented by a ``zoo'' of different pretrained models, together with a selection technique to find the best initialization for the task~\cite{achille2019task2vec}. Better understanding of the distance between tasks may also help characterizing the range in which linearization can be expected to work well. %

Linearization can be made more flexible with architecture changes.
For example, fine-grained classification tasks are typically far from generic models pre-trained on ImageNet. Indeed we observe that LQF drops in performance in fine-grained tasks, which
have been noted to be further from ImageNet \cite{li2020rethinking}.
However, we have shown in  \Cref{sec:fine-grained}, that for this specific case, simply replacing average pooling with bilinear pooling can stretch the performance of LQF in the fine-grained setting.

There are many possible uses for LQF, like any other linear model. Rather than providing an exhaustive list of all possibilities, we showed some illustrative examples. We leave various applications of LQF to future work, including its use in (i) uncertainty quantification, (ii) backward compatibility \cite{shen2020towards}, (iii) leave-one-out cross validation without a validation set \cite{giordano2019swiss}, (iv) active/online learning \cite{strehl2007online}, (v) counterfactual analysis in the first order, (vi) continual learning \cite{Kirkpatrick3521}, (vii) selective forgetting or machine unlearning \cite{golatkar2019eternal,golatkar2020forgetting,guo2019certified}, (viii) training including privacy or fairness constraints \cite{dwork2008differential,kearns2019ethical}, .

\bibliography{main}

\begin{thebibliography}{10}\itemsep=-1pt

\bibitem{achille2019task2vec}
Alessandro {Achille}, Michael {Lam}, Rahul {Tewari}, Avinash {Ravichandran},
  Subhransu {Maji}, Charless {Fowlkes}, Stefano {Soatto}, and Pietro {Perona}.
\newblock {Task2Vec: Task Embedding for Meta-Learning}.
\newblock {\em International Conference on Computer Vision}, 2019.

\bibitem{anonymous2021estimating}
Anonymous.
\newblock Estimating informativeness of samples with smooth unique information.
\newblock In {\em Submitted to International Conference on Learning
  Representations}, 2021.
\newblock under review.

\bibitem{bai2020taylorized}
Yu Bai, Ben Krause, Huan Wang, Caiming Xiong, and Richard Socher.
\newblock Taylorized training: Towards better approximation of neural network
  training at finite width.
\newblock {\em arXiv preprint arXiv:2002.04010}, 2020.

\bibitem{barz2020deep}
Bjorn Barz and Joachim Denzler.
\newblock Deep learning on small datasets without pre-training using cosine
  loss.
\newblock In {\em The IEEE Winter Conference on Applications of Computer
  Vision}, pages 1371--1380, 2020.

\bibitem{basu2020influence}
Samyadeep Basu, Philip Pope, and Soheil Feizi.
\newblock Influence functions in deep learning are fragile.
\newblock {\em arXiv preprint arXiv:2006.14651}, 2020.

\bibitem{csiszar1991least}
Imre Csiszar et~al.
\newblock Why least squares and maximum entropy? an axiomatic approach to
  inference for linear inverse problems.
\newblock {\em The annals of statistics}, 19(4):2032--2066, 1991.

\bibitem{cui2018large}
Yin Cui, Yang Song, Chen Sun, Andrew Howard, and Serge Belongie.
\newblock Large scale fine-grained categorization and domain-specific transfer
  learning.
\newblock In {\em Proceedings of the IEEE conference on computer vision and
  pattern recognition}, pages 4109--4118, 2018.

\bibitem{dwork2008differential}
Cynthia Dwork.
\newblock Differential privacy: A survey of results.
\newblock In {\em International conference on theory and applications of models
  of computation}, pages 1--19. Springer, 2008.

\bibitem{george2018fast}
Thomas George, C{\'e}sar Laurent, Xavier Bouthillier, Nicolas Ballas, and
  Pascal Vincent.
\newblock Fast approximate natural gradient descent in a kronecker factored
  eigenbasis.
\newblock In {\em Advances in Neural Information Processing Systems}, pages
  9550--9560, 2018.

\bibitem{giordano2019swiss}
Ryan Giordano, William Stephenson, Runjing Liu, Michael Jordan, and Tamara
  Broderick.
\newblock A swiss army infinitesimal jackknife.
\newblock In {\em The 22nd International Conference on Artificial Intelligence
  and Statistics}, pages 1139--1147, 2019.

\bibitem{golatkar2019eternal}
Aditya Golatkar, Alessandro Achille, and Stefano Soatto.
\newblock Eternal sunshine of the spotless net: Selective forgetting in deep
  networks.
\newblock {\em Proceedings of the IEEE conference on Computer Vision and
  Pattern Recognition}, 2019.

\bibitem{golatkar2019time}
Aditya Golatkar, Alessandro Achille, and Stefano Soatto.
\newblock Time matters in regularizing deep networks: Weight decay and data
  augmentation affect early learning dynamics, matter little near convergence.
\newblock In {\em Advances in Neural Information Processing Systems}, pages
  10678--10688, 2019.

\bibitem{golatkar2020forgetting}
Aditya Golatkar, Alessandro Achille, and Stefano Soatto.
\newblock Forgetting outside the box: Scrubbing deep networks of information
  accessible from input-output observations.
\newblock {\em European Conference on Computer Vision}, 2020.

\bibitem{golik2013cross}
Pavel Golik, Patrick Doetsch, and Hermann Ney.
\newblock Cross-entropy vs. squared error training: a theoretical and
  experimental comparison.
\newblock In {\em Interspeech}, volume~13, pages 1756--1760, 2013.

\bibitem{griffin2019caltech}
G. Griffin, AD. Holub, and Pietro Perona.
\newblock The caltech 256.

\bibitem{grosse2016kronecker}
Roger Grosse and James Martens.
\newblock A kronecker-factored approximate fisher matrix for convolution
  layers.
\newblock In {\em International Conference on Machine Learning}, pages
  573--582, 2016.

\bibitem{guo2019certified}
Chuan Guo, Tom Goldstein, Awni Hannun, and Laurens van~der Maaten.
\newblock Certified data removal from machine learning models.
\newblock {\em arXiv preprint arXiv:1911.03030}, 2019.

\bibitem{hui2020evaluation}
Like Hui and Mikhail Belkin.
\newblock Evaluation of neural architectures trained with square loss vs
  cross-entropy in classification tasks.
\newblock {\em arXiv preprint arXiv:2006.07322}, 2020.

\bibitem{jacot2018neural}
Arthur Jacot, Franck Gabriel, and Cl{\'e}ment Hongler.
\newblock Neural tangent kernel: Convergence and generalization in neural
  networks.
\newblock In {\em Advances in neural information processing systems}, pages
  8571--8580, 2018.

\bibitem{kearns2019ethical}
Michael Kearns and Aaron Roth.
\newblock {\em The ethical algorithm: The science of socially aware algorithm
  design}.
\newblock Oxford University Press, 2019.

\bibitem{kermany2018identifying}
Daniel~S Kermany, Michael Goldbaum, Wenjia Cai, Carolina~CS Valentim, Huiying
  Liang, Sally~L Baxter, Alex McKeown, Ge Yang, Xiaokang Wu, Fangbing Yan,
  et~al.
\newblock Identifying medical diagnoses and treatable diseases by image-based
  deep learning.
\newblock {\em Cell}, 172(5):1122--1131, 2018.

\bibitem{KhoslaYaoJayadevaprakashFeiFei_FGVC2011}
Aditya Khosla, Nityananda Jayadevaprakash, Bangpeng Yao, and Li Fei-Fei.
\newblock Novel dataset for fine-grained image categorization.
\newblock In {\em First Workshop on Fine-Grained Visual Categorization, IEEE
  Conference on Computer Vision and Pattern Recognition}, Colorado Springs, CO,
  June 2011.

\bibitem{kingma2014adam}
Diederik~P Kingma and Jimmy Ba.
\newblock Adam: A method for stochastic optimization.
\newblock {\em arXiv preprint arXiv:1412.6980}, 2014.

\bibitem{Kirkpatrick3521}
James Kirkpatrick, Razvan Pascanu, Neil Rabinowitz, Joel Veness, Guillaume
  Desjardins, Andrei~A. Rusu, Kieran Milan, John Quan, Tiago Ramalho, Agnieszka
  Grabska-Barwinska, Demis Hassabis, Claudia Clopath, Dharshan Kumaran, and
  Raia Hadsell.
\newblock Overcoming catastrophic forgetting in neural networks.
\newblock {\em Proceedings of the National Academy of Sciences},
  114(13):3521--3526, 2017.

\bibitem{koh2017understanding}
Pang~Wei Koh and Percy Liang.
\newblock Understanding black-box predictions via influence functions.
\newblock {\em arXiv preprint arXiv:1703.04730}, 2017.

\bibitem{kornblith2020whats}
Simon Kornblith, Honglak Lee, Ting Chen, and Mohammad Norouzi.
\newblock What's in a loss function for image classification?
\newblock {\em arXiv preprint arXiv:2010.16402}, 2020.

\bibitem{KrauseStarkDengFei-Fei_3DRR2013}
Jonathan Krause, Michael Stark, Jia Deng, and Li Fei-Fei.
\newblock 3d object representations for fine-grained categorization.
\newblock In {\em 4th International IEEE Workshop on 3D Representation and
  Recognition (3dRR-13)}, Sydney, Australia, 2013.

\bibitem{lee2019wide}
Jaehoon Lee, Lechao Xiao, Samuel Schoenholz, Yasaman Bahri, Roman Novak, Jascha
  Sohl-Dickstein, and Jeffrey Pennington.
\newblock Wide neural networks of any depth evolve as linear models under
  gradient descent.
\newblock In {\em Advances in neural information processing systems}, pages
  8570--8581, 2019.

\bibitem{li2020rethinking}
Hao Li, Pratik Chaudhari, Hao Yang, Michael Lam, Avinash Ravichandran, Rahul
  Bhotika, and Stefano Soatto.
\newblock Rethinking the hyperparameters for fine-tuning.
\newblock {\em arXiv preprint arXiv:2002.11770}, 2020.

\bibitem{li2018explicit}
Xuhong Li, Yves Grandvalet, and Franck Davoine.
\newblock Explicit inductive bias for transfer learning with convolutional
  networks.
\newblock {\em arXiv preprint arXiv:1802.01483}, 2018.

\bibitem{lin2017improved}
Tsung-Yu Lin and Subhransu Maji.
\newblock Improved bilinear pooling with cnns.
\newblock {\em arXiv preprint arXiv:1707.06772}, 2017.

\bibitem{lin2015bilinear}
Tsung-Yu Lin, Aruni RoyChowdhury, and Subhransu Maji.
\newblock Bilinear cnn models for fine-grained visual recognition.
\newblock In {\em Proceedings of the IEEE international conference on computer
  vision}, pages 1449--1457, 2015.

\bibitem{maas2013rectifier}
Andrew~L Maas, Awni~Y Hannun, and Andrew~Y Ng.
\newblock Rectifier nonlinearities improve neural network acoustic models.
\newblock In {\em Proceedings of the International Conference on Machine
  Learning}, volume~30, page~3, 2013.

\bibitem{maji13fine-grained}
S. Maji, J. Kannala, E. Rahtu, M. Blaschko, and A. Vedaldi.
\newblock Fine-grained visual classification of aircraft.
\newblock Technical report, 2013.

\bibitem{martens2015optimizing}
James Martens and Roger Grosse.
\newblock Optimizing neural networks with kronecker-factored approximate
  curvature.
\newblock In {\em International conference on machine learning}, pages
  2408--2417, 2015.

\bibitem{Mu2020Gradients}
Fangzhou Mu, Yingyu Liang, and Yin Li.
\newblock Gradients as features for deep representation learning.
\newblock In {\em International Conference on Learning Representations}, 2020.

\bibitem{nair2010rectified}
Vinod Nair and Geoffrey~E Hinton.
\newblock Rectified linear units improve restricted boltzmann machines.
\newblock In {\em Proceedings of the International Conference on Machine
  Learning}, 2010.

\bibitem{nesterov2013gradient}
Yurii Nesterov.
\newblock {\em Introductory lectures on convex optimization: A basic course},
  volume~87.
\newblock Springer Science \& Business Media, 2013.

\bibitem{nguyen2020leep}
Cuong~V Nguyen, Tal Hassner, Cedric Archambeau, and Matthias Seeger.
\newblock Leep: A new measure to evaluate transferability of learned
  representations.
\newblock {\em arXiv preprint arXiv:2002.12462}, 2020.

\bibitem{Nilsback06}
Maria-Elena Nilsback and Andrew Zisserman.
\newblock A visual vocabulary for flower classification.
\newblock In {\em IEEE Conference on Computer Vision and Pattern Recognition},
  volume~2, pages 1447--1454, 2006.

\bibitem{parkhi12a}
Omkar~M. Parkhi, Andrea Vedaldi, Andrew Zisserman, and C.~V. Jawahar.
\newblock Cats and dogs.
\newblock In {\em IEEE Conference on Computer Vision and Pattern Recognition},
  2012.

\bibitem{quattoni2009recognizing}
Ariadna Quattoni and Antonio Torralba.
\newblock Recognizing indoor scenes.
\newblock In {\em 2009 IEEE Conference on Computer Vision and Pattern
  Recognition}, pages 413--420. IEEE, 2009.

\bibitem{rajaraman2018pre}
Sivaramakrishnan Rajaraman, Sameer~K Antani, Mahdieh Poostchi, Kamolrat
  Silamut, Md~A Hossain, Richard~J Maude, Stefan Jaeger, and George~R Thoma.
\newblock Pre-trained convolutional neural networks as feature extractors
  toward improved malaria parasite detection in thin blood smear images.
\newblock {\em PeerJ}, 6:e4568, 2018.

\bibitem{shen2020towards}
Yantao Shen, Yuanjun Xiong, Wei Xia, and Stefano Soatto.
\newblock Towards backward-compatible representation learning.
\newblock In {\em Proceedings of the IEEE/CVF Conference on Computer Vision and
  Pattern Recognition}, pages 6368--6377, 2020.

\bibitem{strehl2007online}
Alexander Strehl and Michael Littman.
\newblock Online linear regression and its application to model-based
  reinforcement learning.
\newblock {\em Advances in Neural Information Processing Systems},
  20:1417--1424, 2007.

\bibitem{sun2015deeply}
Yi Sun, Xiaogang Wang, and Xiaoou Tang.
\newblock Deeply learned face representations are sparse, selective, and
  robust.
\newblock In {\em Proceedings of the IEEE conference on computer vision and
  pattern recognition}, pages 2892--2900, 2015.

\bibitem{WelinderEtal2010}
P. Welinder, S. Branson, T. Mita, C. Wah, F. Schroff, S. Belongie, and P.
  Perona.
\newblock {Caltech-UCSD Birds 200}.
\newblock Technical Report CNS-TR-2010-001, California Institute of Technology,
  2010.

\end{thebibliography}
\bibliographystyle{ieee_fullname}

\clearpage

\appendix

\section{Supplementary Material for LQF}

In this Supplementary Material we provide additional empirical results (\Cref{sec:additional-results}), more details about the training procedure and implementation used (\Cref{sec:details}), and we derive the expression in \cref{eq:activations-change} for the change in activations when a training sample is removed (\Cref{sec:interpretability-derivation}).

\subsection{Additional results}
\label{sec:additional-results}

\paragraph{Preconditioning with Adam vs.\ K-FAC.} In \Cref{sec:comparison} we train LQF and LQF FC using K-FAC for preconditioning, instead of alternatives like Adam. K-FAC provides several advantages: the learning rate choice is easy to interpret in the case of a quadratic problem (\Cref{sec:kfac}), it easy to analyze theoretically (e.g., in eq.~\ref{eq:adaptive-sgd-time-t}), and it provides better convergence guarantees that Adam. %
Moreover, it provides an approximation of the inverse of the Hessian which we need to compute \cref{eq:activations-change}.
We now test how Adam and K-FAC preconditioning compare to each other in terms of raw test error after hyper-parameter optimization. We train LQF, LQF FC and GaF with Adam on all datasets and report the results in \Cref{tab:adam-errors}. For each dataset we try different learning rates $\eta \in \{0.001, 0.0004, 0.0001\}$, weight decay $\lambda \in \{10^{-5}, 10^{-6}\}$ and augmentation schemes (central crop, random crop, random resized crop) and report the best result. We observe that after hyper-parameter optimization LQF obtains similar final accuracies when trained with Adam and K-FAC (\Cref{tab:adam-errors}). Similarly, for GaF we observe that Adam achieves errors comparable with SGD (in this case without K-FAC). However, K-FAC performs better than Adam when training only the last layer (LQF FC). This suggests that the more sophisticated pre-conditioning of K-FAC is more important to ensure good convergence when optimizing the lower dimensional, badly conditioned problem of LQF FC.

\begin{figure}
    \centering
    \includegraphics[width=.85\linewidth]{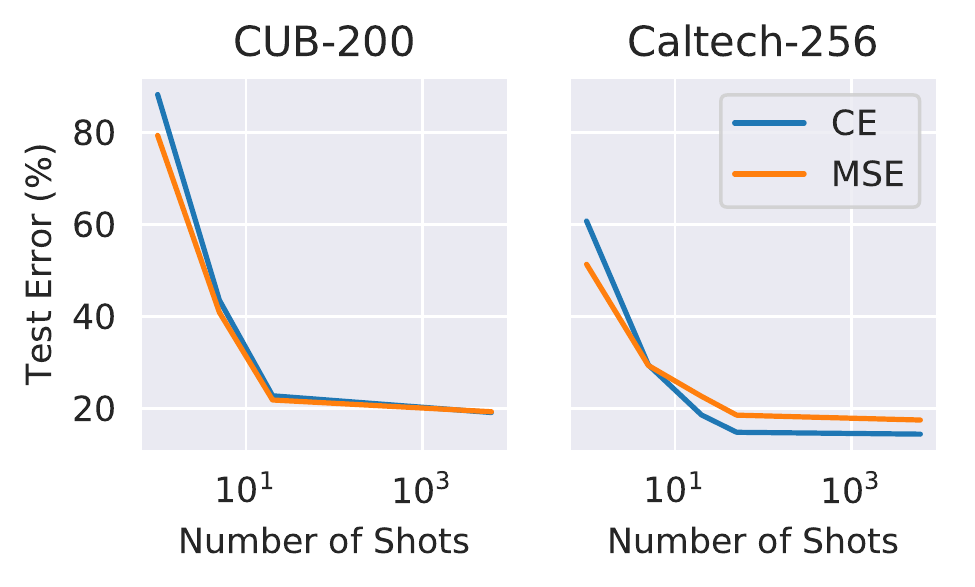}
    \caption{\textbf{Using MSE loss with NLFT.} Plot of the test error obtained by training with NLFT using either cross-entropy of MSE loss for different number of training samples per class (shots). While cross-entropy loss is comparably or better than MSE loss when training with many samples, we observe that in the low-shot regime MSE tends to always outperform CE. This suggests that using MSE loss is not beneficial only for lienarized models.}
    \label{fig:mse-loss}
\end{figure}

\paragraph{MSE loss for non-linear fine-tuning.} In \Cref{fig:mse-loss} we train a standard non-linear network using  cross-entropy loss and MSE loss with different number of training samples. We train with learning rate $\eta \in \{0.1,0.05,0.01,0.001,0.0001\}$ and weight decay $\lambda \in \{0.0001,0.00001\}$ and report the best result. We observe that the MSE loss tends to outperform cross-entropy in the low-data regime, suggesting that some of the benefits of the MSE loss also apply to non-linear fine-tuning.

\begin{table}
    \centering
    \begin{tabular}{lrrr}
    \toprule
     &  LQF &  LQF FC &  GaF \\
    \hline
    \hline
    Caltech-256        & \textbf{14.2} &    17.2 & 16.0 \\
    Chest X-Ray         &  6.6 &     9.6 &  \textbf{6.1} \\
    Malaria Cells &  \textbf{4.2} &     6.1 &  4.8 \\
    MIT-67             & \textbf{20.4} &    25.4 & 23.1 \\
    Oxford Pets              &  \textbf{6.7} &     7.8 &  7.2 \\
    \bottomrule
    \multicolumn{4}{c}{Fine-grained datasets} \\
    \hline
    \hline
    Stanford Dogs      & 12.4 &    13.3 & \textbf{11.7} \\
    Oxford Flowers     &  \textbf{6.5} &    10.8 & 13.8 \\
    CUB-200            & \textbf{23.1} &    29.3 & 28.5 \\
    Aircrafts    & \textbf{33.1} &    45.9 & 45.6 \\
    Stanford Cars      & \textbf{24.0} &    39.2 & 37.0 \\
    \bottomrule
    \end{tabular}

    \caption{\textbf{Test errors using Adam.} We report the test errors obtained by training different linear methods with Adam instead of SGD (in the case of LQF, we also train without K-FAC preconditioning). We note that Adam gives comparable results to SGD+K-FAC for LQF and to standalone SGD for GaF (\Cref{table:dataset-errors}), but is slighly worse than SGD+K-FAC for LQF FC, where we only optimize the last layer.   The only exception where Adam improved results for LQF and GaF is \textit{Chest X-Ray}, possibly due to the more exhaustive hyper-parameter search we used for Adam.
    }
    \label{tab:adam-errors}
\end{table}

\paragraph{Tuning weight-decay.} We did not observe major differences in accuracy by tuning the weight decay parameter suggesting that the inductive bias of linearization is enough to grant good generalization.

\begin{figure}
    \centering
    \includegraphics[width=0.85\linewidth]{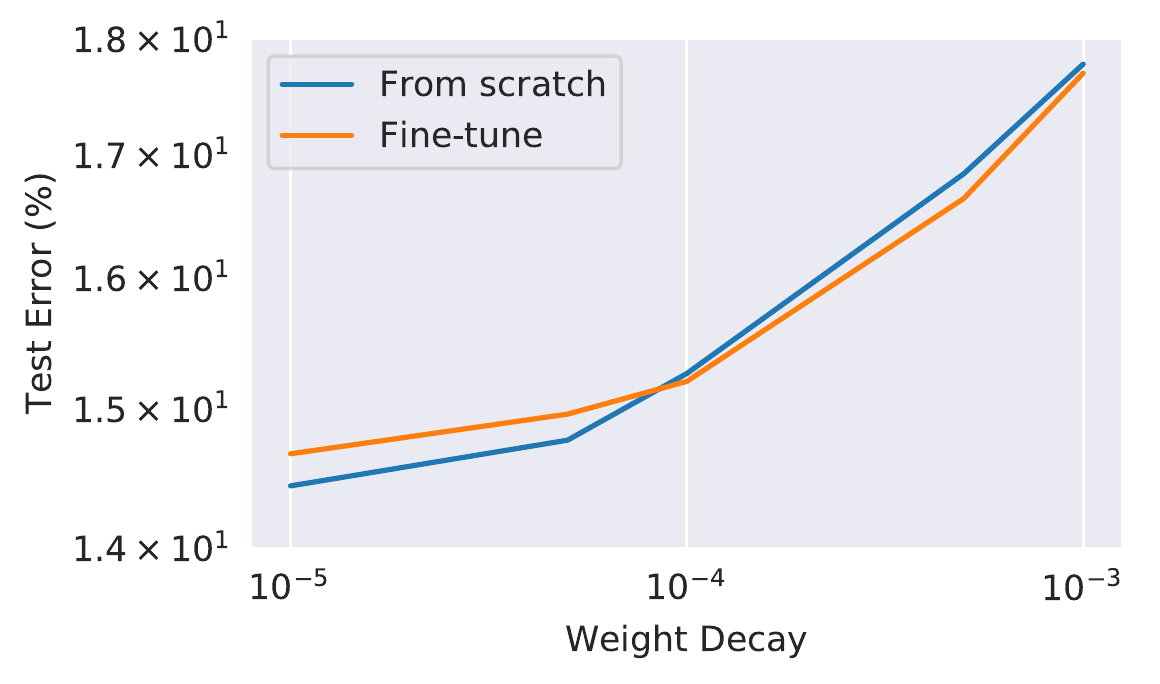}
    \caption{\textbf{Exploring different weight decay values via fine-tuning.} We compare the test error obtained by training from scratch with a given weight decay value $\lambda$, and the test error obtained by loading the solution found with $\lambda_0 = 5 \cdot 10^{-4}$ and fine-tuning with a different value of $\lambda$.  }
    \label{fig:weight-decay-finetuning}
\end{figure}

However, if required, we note that LQF provides a way to tune weight decay efficiently. Since there the LQF loss function is strongly convex, it has a unique global minimum. This implies that, after training with a given value of weight decay, we can increase the value of weight decay and fine-tune the previous solution to converge to the new global minimum. This is in contrast with DNNs, which may remain stuck in a suboptimal local minimum if weight decay is changed after convergence \cite{golatkar2019time}.
In \Cref{fig:weight-decay-finetuning} we show the test accuracy obtained on Caltech-256 when training from scratch with different values $\lambda \in \{10^{-3}, 5 \cdot 10^{-4}, 10^{-4}, 5 \cdot 10^{-5}, 10^{-5}\}$ of weight decay. We then load the solution obtained with $\lambda_0 = 5 \cdot 10^{-5}$ and fine-tune it with different values $\lambda$ of weight decay and compare this with the results obtained training from scratch. We observe that the two approaches (training from scratch and fine-tuning) obtain similar errors, suggesting that indeed for LQF we cab use a cheaper hyper-parameter search based on fine-tuning.

We can also go a step further and automatically optimize weight decay using a validation set. Recall from \cref{eq:mse-optimum} that the weights at convergence as a function of the weight decay $\lambda$ can be written as:
\begin{align}
    w^*(\lambda) = (J^T J + \lambda I)^{-1} J (Y - f_0(X)).
\end{align}
In particular, $w^*(\lambda)$ is a differentiable function of $\lambda$, so we expect to be able to optimize $\lambda$ using gradient descent. In order to do that, consider the validation loss:
\begin{align}
    L_\text{val}^\text{MSE}(w^*(\lambda)) = \frac{1}{|\D_\text{val}|}\sum_{(x,y) \in \D_\text{val}} \| y - f_0(x) - \nabla_w f_0(x) w^*(\lambda) \|^2.
\end{align}
We want to find the value of $\lambda$ that minimizes $L_\text{val}$. The gradient with respect to $\lambda$ is given by
\begin{align}
    \partial_\lambda L_\text{val} = - w^* \cdot \!\!\!\!\! \underbrace{(F + \lambda I)^{-1} \nabla_w L_\text{val}}_{\text{preconditioned gradient of val.~set}}
\end{align}
Note that the second term of $\partial_\lambda L_\text{val}$ is simply the pre-conditioned gradient of the validation loss, which can approximated using K-FAC. We leave further exploration of this research direction to future work.

\subsection{Experimental details}
\label{sec:details}

\paragraph{Choice of datasets.} We have compared the algorithms on a several datasets (\Cref{table:dataset-errors}). Most of the datasets are standard in the fine-tuning or fine-grained classification literature \cite{li2020rethinking,li2018explicit,lin2015bilinear}. We have added additional datasets to this list (\textit{Chest X-ray, Malaria Cells}) so that we could compare on a different domain (medical images instead of natural images). Some of the datasets we use do not have a standard train/test split. In those cases, we use the following splits when reporting the results: For Oxford Flowers we do not merge training and validation data, we only use the training images (1020 samples, instead of the 2040 of train+val). For Caltech-256, we randomly sample 60 images per class to train and test on the remaining images (this scheme is sometimes called Caltech-256-60 in the literature). For Malaria Cell, we randomly select 75\% of samples for training, and test of the remaining ones.

\paragraph{Pre-training.} In all our experiments we use a ResNet-50 backbone pretrained on ImageNet using SGD (we use the reference PyTorch pretraining scheme: 90 epochs with cross-entropy loss, learning rate 0.1 decayed by a factor 10 every 30 epochs, momentum 0.9). We use Leaky ReLU activations for LQF, while we use standard ReLU activations for all other methods. To ensure that the comparison is as fair as possible, we obtain the weights of the Leaky ReLU network by starting from the pretrained ReLU network, changing the activations, and then fine-tune on ImageNet with SGD for another epoch.\footnote{Interestingly, we observe that even without fine-tuning the weights learned for ReLU activations still achieve a good test error when used with Leaky ReLU activations. This makes fine-tuning with the new activations particularly easy.} We also tried training the backbone from scratch on ImageNet using Leaky ReLU activations, but did not observe any major difference in performance between the two.

\paragraph{Optimization.} In all experiments, we train using SGD with momentum 0.9 and batch size 28. We search for the learning rate in $\eta \in \{0.01, 0.001\}$ and weight decay $\lambda \in \{10^{-4}, 10^{-5}\}$. We report the best result. Instead of applying weight decay directly to the gradient update as often done, we add the $\ell_2$ weight penalty to the loss function. This is required when using K-FAC to compute the correct pre-conditioned update, but it does not otherwise change the udpate equation when not using K-FAC.

In \Cref{sec:additional-results} we present additional results using Adam. In this case we search for the learning rate in $\eta \in \{0.001, 0.0004, 0.0001\}$ and weight decay $\lambda \in \{10^{-5}, 10^{-6}\}$. When training GaF with Adam we use $\beta_1 = 0.5$ as suggested in \cite{Mu2020Gradients}, otherwise we use $\beta_1 = 0.9$.

\paragraph{Data augmentation.} Since different tasks may require different types of data augmentation, we train with different augmentation schemes (center crop, random crop, resized random crop) and report the best result. For \textit{center crop}, we resize the image to $256 \times 256$ and extract the central crop of size $224 \times 224$, while for \textit{random crop} we extract a $224 \times 224$ in a random position and also apply a random horizontal flip. In both cases we test with center crops. We also try the \textit{resized random crops} augmentation  commonly used for ImageNet pre-training. We found this augmentation to be too strong for some tasks, and it gives significantly better results only in the case of \textit{Chest X-Rays}. For this reason, and to save computation time, we only train random crop with one combination of hyper-parameters (learning rate $\eta=0.001$ and weight decay $\lambda=10^{-5}$).

\paragraph{Linearization.} We follow the procedure of \cite{Mu2020Gradients} to linearize the network. The main difference is that we do not fuse the batch norm layer with the previous convolutional layer, but rather we also linearize the batch norm layer. We did not observe major difference in performance by using one or the other solution, but we opted for the latter to keep the structure of the linearized and original network as close as possible when performing the comparison. In order to ensure linearity, for all linear models (LQF, LQF FC, GaF, Standard FC) we put batch normalization in eval mode when training (that is, we use the (frozen) running mean and variance to whithen the features rather than using the current mini-batch statistics).

\paragraph{B-LQF.} The only change necessary for B-LQF with respect to stnadard LQF is to replace the global average pooling before the classification layer with a linearized version of bilinear pooling with square root normalization \cite{lin2017improved}.
Let $z \in \R^{HW \times C}$ be the last convolutional layer features, where $H$ and $W$ are the spatial dimensions and $C$ is the number of channels. Their covariance matrix $\Sigma \in \R^{C\times C}$ can be written as:
\begin{align}
    \Sigma = \frac{1}{N} z^T (I - \frac{1}{N} \mathbbm{1}) z = z^T A z
\end{align}
where $I$ denotes the identity matrix and $\mathbbm{1}$ denotes the matrix of all ones and $N = H \cdot W$. Using the same notation as \cite{Mu2020Gradients}, let $h(z(r)) = \sqrt{\Sigma} =  \sqrt{z^T(r) A z(r)}$. To compute the linearization forward pass of the layer we need to compute:
\begin{align}
    \partial_r h(z(r)) = \frac{1}{2} \sqrt{\Sigma}^{-1} (\partial_r z^T A z + z^T A \partial_r z).
\end{align}
Note that we already have $\sqrt{\Sigma}$ from the forward pass of the base model, so computing the forward pass of the linearized model does not add much to the complexity.

\paragraph{Robustness to optimization hyper-parameters.}
As mentioned in \Cref{sec:robustness}, to obtain the plot in \Cref{fig:comparison} (right) we train LQF and NLFT with SGD with $\eta \in \{0.05,0.01,0.005,0.001,0.0001\}$, batch size $b \in \{16, 32, 64\}$ and weight decay $\lambda \in \{10^{-5},10^{-4},5 \cdot 10^{-4}\}$ and we report the best result. Due to the larger search space, we report the results only for a representative subset of the datasets: \textit{Oxford Flowers, MIT-67,CUB-200, Caltech-256, Stanford Dogs, FGVC Aircrafts}.

\pagebreak

\subsection{Derivation of interpretability (eq.~\ref{eq:activations-change})}
\label{sec:interpretability-derivation}

Recall that in the case of LQF the hessian of the MSE loss (\cref{eq:mse-loss}) is given by $H = F + \lambda I$ where $F = \frac{1}{N} J J^T = \frac{1}{N} \sum_{j=1}^N g_j^T g_j $, where $g_j = \nabla_w f_0(x_j)$ are the Jacobian of the $i$-th sample computed at the linearization point $w_0$. To keep the notation simple, assume this a binary classification problem, so that the label $y \in \{0, 1\}$ is a scalar and the Jacobian $g_i$ is a row vector (a similar derivation holds for a multi-class problem).

Let $w^*$ be the optimum of the loss function computed using $N$ training samples.
Since in a quadratic problem a Newton-update converges to the optimum in a single step, we can write the new optimal weights obtained after removing the $i$-th training sample -- that is, training with only $N-1$ samples -- as:
\begin{align}
    w_{-i}^* = w^* - H_{-i}^{-1} \nabla_w L_{-i}(w^*)
    \label{eq:newtown-step}
\end{align}
where  $L_{-i}$ and $H_{-i}$  are respectively the training loss computed without the sample $i$ and $H_{-i}$ is its hessian. Note that we can write $L_{-i}(w)$ as
\[
L_{-i}(w) = L(w) - \frac{1}{N} \|y_i - f_0(x_i) - g_i w\|^2.
\]
Since $w^*$ is the optimum of the original problem, we have $\nabla L(w^*) = 0$. Using this and the previous equation we get:
\[
\nabla_w L_{-i}(w^*) =  \nabla_w L(w^*) - g_i^T e_i = - g_i^T e_i
\]
where $e_i \equiv \frac{1}{N}(y - g_i w^*) \cdot $ is the (weighted) prediction error on the sample $x_i$.
Plugging this in \cref{eq:newtown-step} we have
\[
w_{-i}^* = w^* + H_{-i}^{-1} g_i^T e_i.
\]
We now derive an expression for $H_{-i}^{-1}$ as a function of $F$. Note that
\[
H_{-i} = F_{-i} + \lambda I = \frac{N}{N-1} F + \lambda I - \frac{1}{N-1} g_i^T g_i.
\]
Let $A = \frac{N}{N-1} F + \lambda I$. Note that $H_{-i}$ is a rank-1 update of $A$. Using the Sherman–Morrison formula for the inverse of a rank-1 update (or, more generally, the Woodbury identity in the case of a multi-class problem), we have
\[
H_{-i}^{-1} = (A - \frac{1}{N-1} g_i^T g_i)^{-1} = A^{-1} - \frac{A^{-1} g_i^T g_i A^{-1}}{N - 1 + g_iA^{-1}g_i^T}
\]
The activation change $\Delta f(x_\test) = f_{w^*}(x_\test) - f_{w^*_{-1}}(x_\test)$ after removing a training sample is then given by:
\begin{align}
\Delta f(x_\test) &= g_\test (w^* - w^*_{-i})\nonumber\\
&= g_\test H_{-i}^{-1} g_i^T e_i \nonumber\\
&=e_i g_\test A^{-1} g_i -  \frac{e_i g_\test A^{-1} g_i^T g_i A^{-1} g_i}{N - 1 + g_i A^{-1}g_i^T}\nonumber
\end{align}
Reorganizing the terms, we obtain the following expression for \cref{eq:activations-change} in the main paper:
\begin{align}
\boxed{\Delta f(x_\test)  =(1 - \frac{\alpha}{N-1 + \alpha}) e_i g_\test A^{-1} g_i^T}
\label{eq:activation-change-correct}
\nonumber
\end{align}
where we defined $\alpha = g_i A^{-1} g_i^T$.
In particular, note that for large datasets ($N \gg 1$) the change in activations is simply given by $e_i g_\test A^{-1} g_i^T$ which measures the similarity of the Jacobian of the sample $x_\test$ and the jacobian of $x_i$ under the metric induced by the kernel $A^{-1}$. Finally, note that $A^{-1}$ is the inverse of the Fisher Information Matrix $F$ plus a multiple of the identity, which we can easily estimate using K-FAC \cite{martens2015optimizing}. This is particularly convenient since we are already computing the K-FAC approximation to pre-condition the gradients.

In the case of a multiclass problem, the Jacobian $g_i$ is a $C \times D$ matrix, where $C$ is the number of classes and $D$ is the number of parameters. In this case, we get a similar expression:
\[H_{-i}^{-1} = A^{-1} - A^{-1}g_i^T((N-1) I_{c} + g_iA^{-1}g_i^T)^{-1}g_iA^{-1}\]
The activation change $\Delta f(x_\test) = f_{w^*}(x_\test) - f_{w^*_{-1}}(x_\test)$ after removing a training sample is then given by:
\begin{align}
\Delta f(x_\test) =& g_\test (w^* - w^*_{-i})\nonumber\\
=& g_\test H_{-i}^{-1} g_i^T e_i \nonumber\\
=& g_\test A^{-1} g_i^T e_i \nonumber\\
& - g_\test A^{-1}g_i^T((N-1) I_{c} + g_iA^{-1}g_i^T)^{-1}g_iA^{-1}g_i^T e_i \nonumber
\end{align}
Reorganizing the terms, we have:
\[
\Delta f(x_\test) = g_\test A^{-1} g_i^T (I_C \\
- M^{-1} \alpha) e_i
\]
where $M=(N-1) I_{c} + \alpha$, $\alpha = g_i A^{-1} g_i^T$ is now a $C \times C$ matrix and $I_C$ denotes the $C \times C$ identity.
\end{document}